\documentclass[manuscript,screen]{acmart}
\usepackage{amsmath,amsfonts}
\usepackage{graphicx}%
\usepackage{array}
\usepackage{algorithm}
\usepackage{algorithmicx}
\usepackage{algpseudocode}
\usepackage{subcaption}

\setcopyright{none} 
\renewcommand\footnotetextcopyrightpermission[1]{} 
\AtBeginDocument{%
  }

\acmISBN{978-1-4503-XXXX-X/2018/06}

\begin{document}

\title{Optimizing Graph Causal Classification Models: Estimating Causal Effects and Addressing Confounders}

\author{Simi Job}
\affiliation{%
  \institution{School of Mathematics, Physics, and Computing, University of Southern Queensland}
  \country{Australia}}
\email{simi.job@unisq.edu.au}

\author{Xiaohui Tao}
\affiliation{%
  \institution{School of Mathematics, Physics, and Computing, University of Southern Queensland}
  \country{Australia}}
\email{Xiaohui.Tao@unisq.edu.au}

\author{Taotao Cai}
\affiliation{%
  \institution{School of Mathematics, Physics, and Computing, University of Southern Queensland}
  \country{Australia}}
\email{Taotao.Cai@unisq.edu.au}

\author{Haoran Xie}
\affiliation{%
  \institution{School of Data Science, Lingnan University}
  \city{Tuen Mun}
  \country{Hong Kong}}
\email{hrxie@ln.edu.hk}

\author{Jianming Yong}
\affiliation{%
  \institution{School of Business, University of Southern Queensland}
  \country{Australia}}
\email{jianming.yong@unisq.edu.au}

\author{Xin Wang}
\affiliation{%
  \institution{Schulich School of Engineering, University of Calgary}
  \city{Calgary}
  \country{Canada}}
\email{xcwang@ucalgary.ca}

\renewcommand{\shortauthors}{Job et al.}

\begin{abstract}
 Graph data is becoming increasingly prevalent due to the growing demand for relational insights in AI across various domains. Organizations regularly use graph data to solve complex problems involving relationships and connections. Causal learning is especially important in this context, since it helps to understand cause-effect relationships rather than mere associations. Since many real-world systems are inherently causal, graphs can efficiently model these systems. However, traditional graph machine learning methods including graph neural networks (GNNs), rely on correlations and are sensitive to spurious patterns and distribution changes. On the other hand, causal models enable robust predictions by isolating true causal factors, thus making them more stable under such shifts. Causal learning also helps in identifying and adjusting for confounders, ensuring that predictions reflect true causal relationships and remain accurate even under interventions. To address these challenges and build models that are robust and causally informed, we propose \textit{CCAGNN}, a Confounder-Aware causal GNN framework that incorporates causal reasoning into graph learning, supporting counterfactual reasoning and providing reliable predictions in real-world settings. Comprehensive experiments on six publicly available datasets from diverse domains show that \textit{CCAGNN} consistently outperforms leading state-of-the-art models.
\end{abstract}




\maketitle

\section{Introduction}
In real-world applications, graphs are often used to represent complex relationships between variables in a system, which allows for the modeling of potential causal interactions. However, these graph-based models face several challenges such as confounding factors, structural changes and incompleteness. To address these issues, it is important to study modifications to the graph structure, such as adjusting node values or altering connections. Exploring these variations helps us to understand their impact on classification accuracy, model robustness and prediction reliability. 

\begin{figure}[h] 
\centering

\includegraphics[width=0.80\textwidth]{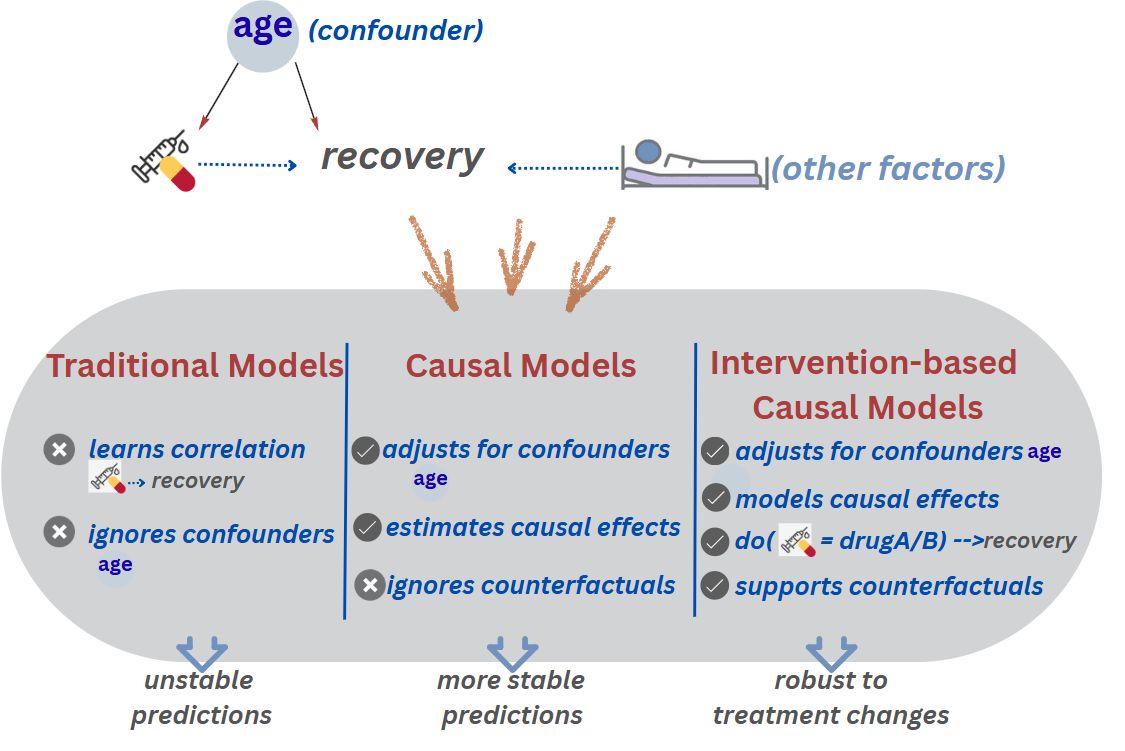}
\caption{Traditional vs causal GNN in predicting patient recovery; causal GNN isolates medication's true effect by adjusting for confounders (age, other patient factors); intervention-based causal GNN models treatment changes for robust predictions.}
\label{fig:motivation}
\end{figure}

Modeling and predicting outcomes in complex systems such as healthcare, marketing and finance, plays an important role in decision making. By uncovering causal relationships, these models reveal how different factors are connected, helping to make better decisions about treatments, marketing strategies and investments. However, confounding factors make it challenging to understand true causal effects. For example, in medicine, confounding factors such as a patient's age or pre-existing conditions can bias treatment outcome analysis if they are not controlled. Similarly, in marketing, seasonality is a main confounding factor impacting promotional strategies. Likewise, economic conditions can confound the impact of financial strategies in investments. Fig.~\ref{fig:motivation} illustrates the differences between traditional, causal and intervention-based causal models in predicting patient recovery in a clinical setting. In traditional models, predictions are based on correlations, which can be misleading due to confounding factors such as age that influence both treatment and recovery. For instance, dosages may differ between younger and older patients due to metabolism and risk factors, and older patients often have weaker immune systems and slower healing rates. Causal models improve accuracy by adjusting for these confounders to isolate the true effect of medication. Intervention-based causal models extend this further by simulating treatment changes, enabling more robust and reliable predictions under different settings.

This research aims to address these challenges by developing novel causal classification models using graph neural networks (GNNs) that optimise graph representations through feature disentanglement and mutual-information based causal learning, thereby enhancing the reliability of classification models. Causality-based classification has been explored by many researchers including \cite{zhao2024twist, wu2024graph}. The causal classification approach in the ICL framework \cite{zhao2024twist} employs disentanglement; however, explicit separation of causal effects requires counterfactual reasoning, leading to more reliable identification of causal features, an aspect addressed by our framework. Similarly, CaNet \cite{wu2024graph} separates independent latent factors in the causality domain, but our work enhances the extraction of reliable causal features through mutual information based causal learning. The proposed Causal Confounder-Aware Graph Neural Network (CCAGNN) framework improves graph classification by disentangling causal and non-causal features through counterfactual interventions and mutual information-based contrastive training, enabling more effective modeling of causal relationships in graph data. This research makes the following key contributions:

\begin{itemize}
    \item Attention-Guided Feature Disentanglement: Introduces an attention-guided feature disentanglement mechanism using confidence-supervised gating to separate and control causal and non-causal features in GNNs.
    \item Mutual information-based causal feature learning: A multi-branch mutual information framework leveraging conditional and prediction-relevant mutual information losses to enhance causal feature learning and disentanglement.
    \item Latent Confounder-Aware Feature Fusion: Presents a representation fusion strategy guided by gating and feature decorrelation to mitigate the influence of latent confounding factors and improve classification robustness.
\end{itemize}

\section{Related Work}
In this section, we review works related to graph neural networks and causality.

\subsection{Graph Neural Networks}

Graph Neural Networks have been used for graph-based learning in several domains such as recommendation systems \cite{wang2024graph, liu2024selfgnn}, temporal forecasting \cite{kong2024spatio} and anomaly \cite{zhou2024reconstructed} and fraud \cite{reddy2025understanding} detection in applications including traffic \cite{kong2024spatio}, IoT \cite{zhou2024reconstructed, sharma2024image}, healthcare \cite{manivannan2024graph, xu2024predicting}, finance \cite{reddy2025understanding} etc. In SelfGNN \cite{liu2024selfgnn}, GNNs are employed to model short-term user behaviour graphs based on time intervals, capturing collaborative patterns among user interactions to enhance the recommendation process. In a similar domain, but focusing on multi-modal data, GNNs are used in GNNMR \cite{li2024graph} to learn uni-modal user-item embeddings from separate bipartite graphs for each modality using mutual knowledge distillation to integrate different modalities. In IoMT, GNNs have been used to enhance image and video analysis by extracting highly discriminative features through modeling visual data as graphs \cite{sharma2024image}. GNNs have also been used in healthcare decision making with graph convolutional neural networks applied to predict ICU interventions by modeling patient time series data as graphs \cite{xu2024predicting}. In the same domain, GNNs were used to optimize resource allocation by modeling healthcare networks as graphs to support resource distribution forecasting through spatio-temporal analysis \cite{manivannan2024graph}. In spatio-temporal analysis, GNNs have also been used for traffic forecasting by modeling heterogeneous spatio-temporal graphs, capturing high-order relationships through prior and posterior message passing for enhanced forecasting \cite{ju2024cool}.

\subsection{Causality and Graph Neural Networks}
Causality has been introduced into GNN frameworks for modeling causal relationships in complex graph systems, providing significant performance improvements in domains such as healthcare \cite{liu2024cignn}, recommendation systems, fault diagnosis \cite{liu2024causal}, demand forecasting \cite{miraki2024electricity} etc. Liu et al. \cite{liu2024causal} developed the Causal Disentangled GNN (CDGNN) to effectively disentangle causal and bias features in industrial fault diagnosis. It uses an attention mechanism for a stable, causality-based system while separating relevant fault signals from irrelevant noise.  However, the framework involves a repetitive process of generating subgraphs and embeddings. On the other hand, Sui et al. \cite{sui2024invariant} introduced the Invariant Graph Learning (IGL) approach for causal effect estimation across multiple environments. It learns invariant confounders and spillover effects, handling distribution shifts, although the environment generation adds complexity to the framework. Liu et al. \cite{liu2024cignn} proposed CiGNN, where a spatio-temporal GNN is used to improve cuffless blood pressure estimation by combining spatial information from the causal graph and temporal data from cardiac signals. Causality with GNNs was also used in recommendation systems through Neural Causal Graph Collaborative Filtering (NCGCF) framework \cite{wang2024neural}, where causal graph-based embeddings capture complex dependencies to enhance collaborative filtering models. The noise in user behaviours when modeling complex user-item relationships in recommender systems is addressed using a GNN-based Causal Denoising Framework (GCDF) \cite{zhao2024causal}, which filters out noisy connections based on causal relationships. In the Knowledge Graph Denoising-based Causal Recommendation (KGDCR) framework \cite{guo2024causal}, semantic information from the knowledge graph is aggregated using Graph Attention Networks to capture personalized user preferences thereby improving recommendation performance. 

Causality has also been integrated with GNNs in demand forecasting through eXplainable Causal Graph Neural Network (X-CGNN) \cite{miraki2024electricity}, which employs GNNs to model interdependencies in multivariate electricity demand while providing causal and interpretable explanations for its predictions. In the domain of load forecasting, \cite{huang2025causality} proposed Causality-Aware Dynamic Multi-Graph Convolutional Network (CADGN) framework that employs GNNs to model causal relationships and dynamically evolving critical nodes, enabling more accurate spatio-temporal predictions for electric vehicle charging station loads.

\section{Preliminaries}

\subsection{Graph Attention Networks}

Graph Attention Networks(GAT) \cite{velivckovic2017graph} are a type of graph neural network that incorporate the attention mechanism. They learn to assign different weights called attention coefficients to neighbouring nodes, enabling each node to compute a weighted sum of its neighbours' features. The graph attention layer updates node features as shown in Eq.~\ref{eq-Prelim1} \cite{alma991006690901904691}, where the attention coefficients $ \alpha_{ij} $ are computed as shown in Eq.~\ref{eq-Prelim2} \cite{alma991006690901904691}. Here, $W$ denotes a learnable linear transformation, $a$ the attention vector and $\sigma$, a nonlinear activation, and the neighbourhood $\mathcal{N}(i)$ includes node $i$ and its immediate neighbours.

\begin{equation} \label{eq-Prelim1}
H_i^{k} = \sigma \left( \sum_{j \in \mathcal{N}(i)} \alpha_{ij} W H_j^{k-1} \right),
\end{equation}

\begin{equation} \label{eq-Prelim2}
\alpha_{ij} = \frac{
\exp\left(\text{LeakyReLU}\left( \mathbf{a}^\top \left[ W H_i^{k-1} \parallel W H_j^{k-1} \right] \right)\right)
}{
\sum_{l \in \mathcal{N}(i)} \exp\left(\text{LeakyReLU}\left( \mathbf{a}^\top \left[ W H_i^{k-1} \parallel W H_l^{k-1} \right] \right)\right)
},
\end{equation}

\subsection{Causal Learning}

\begin{figure}[h]
\centering
\includegraphics[width=0.65\textwidth]{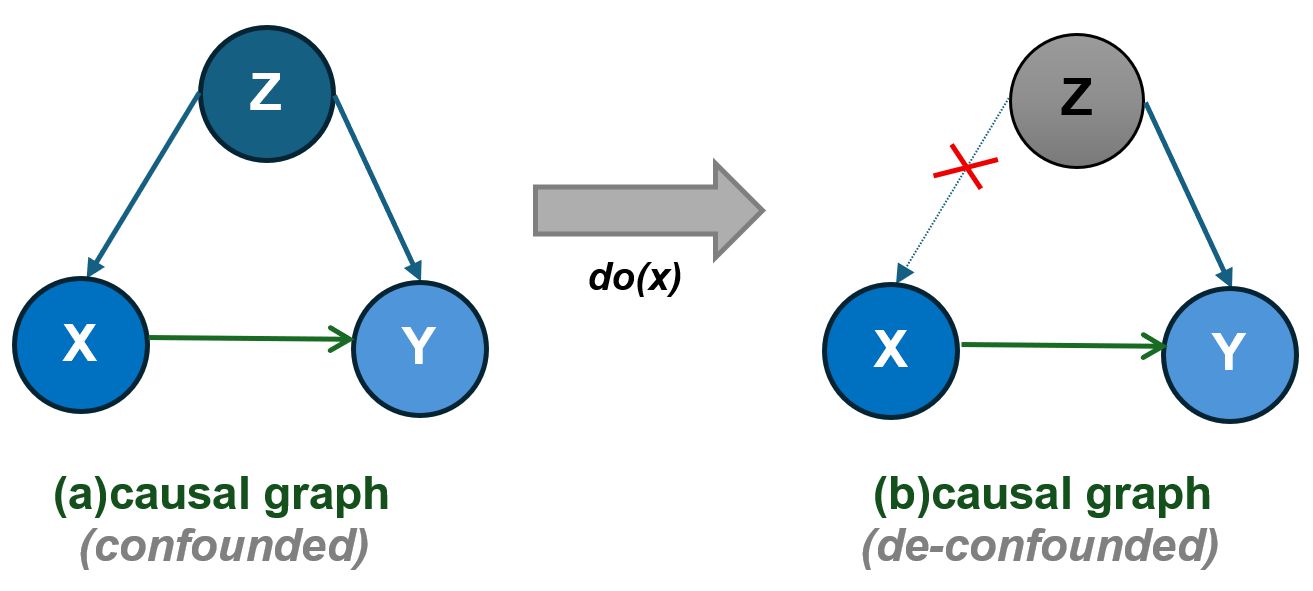}
\caption{Causal graphs: (a) \textit{Z} influences both \textit{X} and \textit{Y}; (b) Isolates true effect of \textit{X} on \textit{Y}, independent of \textit{Z}}
\label{fig:causal}
\end{figure}
Causal Learning is the process of inferring cause and effect relationships from data, as opposed to traditional approaches that focus on correlations. This process can be formally defined in terms of do-calculus, which provides a solid framework for identifying and estimating causal effects through structural manipulations under diverse conditions \cite{pearl2012calculus}. Do-calculus consists of a set of formal rules that provide a structured approach to estimate causal effects and draw valid causal conclusions from data. This involves the concept of \textit{interventions}, represented by the operator \textit{do(X)}, which corresponds to actively setting the variable \textit{X} to a specific value, effectively eliminating the influence of its prior causes. Using the same example from Fig.~\ref{fig:motivation}, Fig.~\ref{fig:causal} illustrates causal graphs depicting a treatment-outcome relationship. Subfigure (a) shows an observational setting, where confounders (\textit{Z}) such as \textit{age}, influence the treatment \textit{X}(\textit{drug type}) and the outcome \textit{Y} (\textit{recovery}). However, without adjusting for \textit{Z}, we cannot determine whether the recovery is caused by the treatment or by the confounding effect of age. In subfigure (b), an intervention is applied by fixing the treatment to \textit{drug A} using \textit{do(X = drug A)}, thus breaking the link from \textit{Z}, thereby removing the influence of confounders. This enables the estimation of the true causal effect of \textit{X} on \textit{Y}, independent of \textit{Z} \cite{jiao2024causal, GopnikAlison2007CLPP}.

\section{Methodology}
This section presents the problem formulation and introduces the CCAGNN framework for developing a confounder-aware causal graph neural network.

\subsection{Problem Definition}
Given a graph $G = (V,E)$, where V denotes the set of nodes and E the set of edges, the primary objective is to predict the class label $y{_{i}}$ for each node $v{_i} \in V$. The intervention on a subset of nodes $X{_{intervened}} \subseteq V$ modifies node features to produce an intervened graph $G'$ with updated features $X'$: $X'=Intervene(X,X{_{intervened}})$, where $Intervene(.)$ denotes the controlled modification of node features for the intervened nodes. This enables to analyse how interventions on node features causally influence node classification, supporting robust predictions.

\begin{figure*}[h]
    \centering
    \begin{subfigure}[b]{0.40\textwidth}
        \centering
        \includegraphics[width=\textwidth]{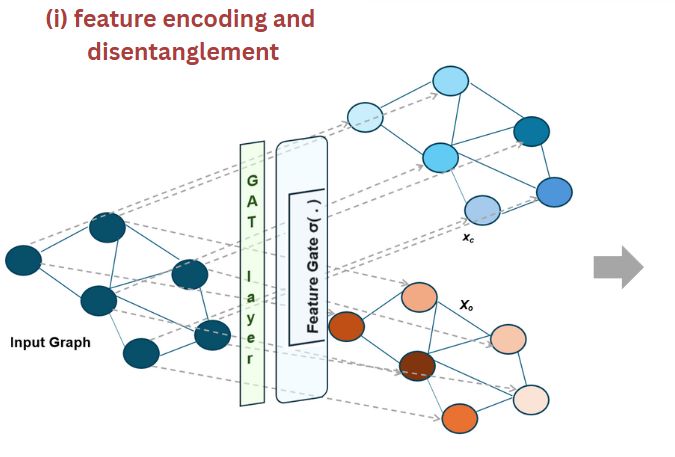}

        \label{fig:design1}
    \end{subfigure}
    \hfill
    \begin{subfigure}[b]{0.58\textwidth}
        \centering
        \includegraphics[width=\textwidth]{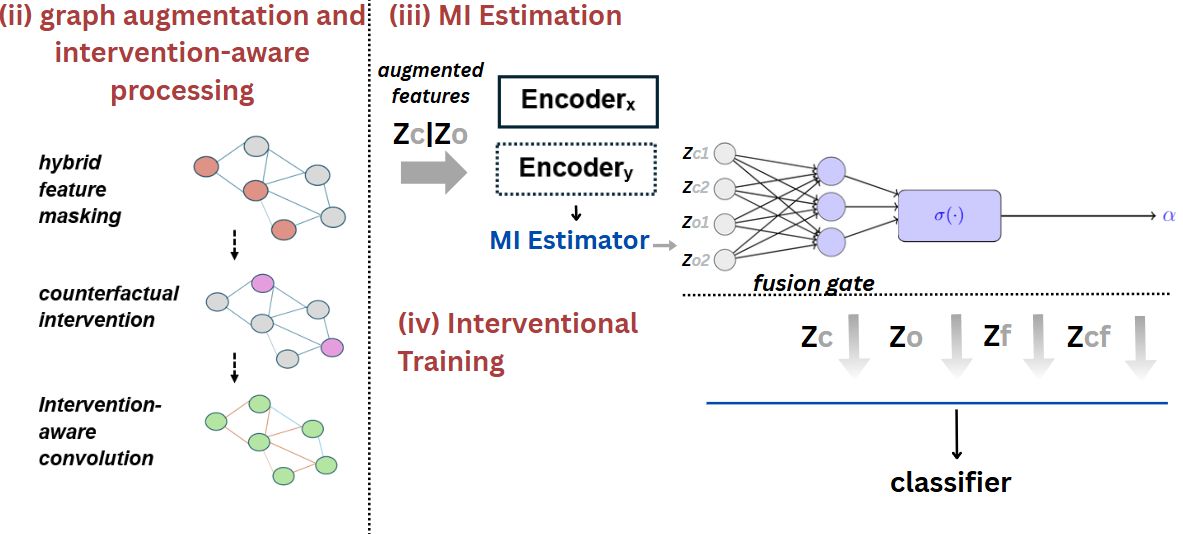}

        \label{fig:design2}
    \end{subfigure}
    \caption{CCAGNN Architecture: (i) Node features are split into causal and non-causal parts using gated GAT layers for structure-aware encoding; (ii) Hybrid feature masking, attention-guided noise and structural changes simulate interventions, processed through dual GAT pipelines for causal and non-causal features; (iii) A dual-encoder estimates and minimizes mutual information between causal and non-causal features with a confidence-based fusion gate for disentangled representations; (iv) Counterfactual interventions and multi-branch learning with gated fusion allows robust predictions.}
    \label{fig:Design}
\end{figure*}

\subsection{Causal Confounder-Aware Graph Neural Networks (CCAGNN)}

In this section, we detail the architecture of the Causal Confounder-Aware Graph Neural Network model. Fig.~\ref{fig:Design} illustrates the model's overall architecture and Algorithm~\ref{alg:CCAGNNalg} details the procedure for \textit{CCAGNN}. 

\begin{algorithm}[H]
\caption{CCAGNN: Causal  Confounder-Aware Graph Neural Networks} \label{alg:CCAGNNalg}
\begin{algorithmic}[1]
\Require Node features \( X \in \mathbb{R}^{N \times d} \), edge index \( E \), labels \( Y \)

\State Representation Learning:
\State \quad Compute node embeddings: \( H = \text{GNN}(X, E) \)

\State Disentanglement:
\State \quad Learn gate \( \alpha = \sigma(f(H)) \)
\State \quad Compute \( X_c = \alpha \cdot H \) \Comment{\textit{causal part}}
\State \quad Compute \( X_o = (1 - \alpha) \cdot H \) \Comment{\textit{non-causal part}}

\State Interventions:
\State \quad Generate counterfactual embeddings \( X_c^{cf}, X_o^{cf} \)

\State Mutual Information Estimation:
\State \quad Estimate \( \mathcal{I}_{\text{inv}}, \mathcal{I}_{\text{pred}}, \mathcal{I}_{\text{cond}} \) \Comment{\textit{invariance, prediction-related, conditional}}

\State Classification:
\State \quad Fuse embeddings: \( Z = \alpha \cdot X_c + (1 - \alpha) \cdot X_o \)
\State \quad Predict labels from \( Z \)

\State Loss Composition:
\State \quad Combine supervised and regularization losses to get \( \mathcal{L}_{\text{total}} \)

\State \Return \( \mathcal{L}_{\text{total}} \)
\end{algorithmic}
\end{algorithm}

We propose Causal Confounder-Aware Graph Neural Networks (CCAGNN), a novel framework for graph classification that models node interactions from a causal perspective. In graph data, a node's label is expected to be influenced by both contextual and object-specific features. To capture these causal dependencies, we design a causally aware architecture that disentangles and models these two feature types, following \cite{job2025hebcgnn} and \cite{sui2022causal}. To further uncover causal relationships, we apply counterfactual reasoning by generating predictions through targeted interventions on these components. These interventions enable the model to simulate how changes in these features impact predictions, thereby enhancing causal understanding. By decoupling context-object pairs during interventions, the model maintains appropriate uncertainty and avoids overly deterministic outcomes under varying conditions. The causal learning process is embedded in the GNN architecture through attention-based encoding, interventional message passing and contrastive training, with the latter implemented through a mutual information regularization mechanism between the context and object representations. Algorithm~\ref{alg:CCAGNNalg} details the procedure for \textit{CCAGNN}. The algorithm starts by taking the input features, and lines 1-2 involve representation learning where the GNN processes the features to learn node representations. In the next step (lines 3-6), disentanglement occurs, where a gating vector $\alpha$ is learned using a sigmoid function over the transformed embeddings, followed by the computation of causal and non-causal parts. Lines 7-8 involve the generation of counterfactual causal and non-causal embeddings for intervention analysis. Mutual information estimation follows (lines 9-10) to guide disentanglement. In the classification stage (lines 11-13), reconstruction of the final embedding \textit{Z} occurs, based on which node label predictions are made. Lines 14-16 involve loss computation, combining supervised and regularization terms. These stages are discussed in detail in the following subsections.

\subsubsection{Feature Encoding} 
In the first stage, raw node features are processed using GATv2-based \cite{brody2021attentive} graph convolutions to generate informative node representations. These embeddings are separated into causal and non-causal components through a learnable sigmoid-based gating mechanism. For each input feature vector $\mathbf{x} \in \mathbb{R}^d$, the disaggregation into causal and non-causal components is performed as shown in Eq.~\ref{eq-Feature1}, where $\mathbf{g} \in [0,1]^d $ denotes the learnable feature-wise gate, $\mathbf{W}_g$ and $\mathbf{b}_g$ are the learnable weight and bias, and $\mathbf{x}_c$, $\mathbf{x}_o$ represent the causal and non-causal features respectively. This disentanglement enables the model to isolate features that are critical for causal reasoning. Dedicated GATv2 convolutional layers are applied to $\mathbf{x}_c$ and $\mathbf{x}_o$ for further refinement, while maintaining their respective structural dependencies.

\begin{equation}\label{eq-Feature1}
\begin{aligned}
\mathbf{g} &= \sigma(\mathbf{W}_g \mathbf{x} + \mathbf{b}_g) \in [0, 1]^d \\
\mathbf{x}_c &= \mathbf{g} \odot \mathbf{x} \\
\mathbf{x}_o &= (1 - \mathbf{g}) \odot \mathbf{x}
\end{aligned}
\end{equation}

\subsubsection{Graph Augmentation and Intervention-Aware Processing}
To present the model with varied causal contexts, a causality-aware data augmentation module introduces controlled variations to the node features and graph structure. This module leverages Graph Attention Networks to effectively learn from these causal interventions. The augmentations include: i) Attention-guided noise injection: Noise is dynamically scaled by attention scores from GAT layers, targeting variations at the most influential nodes. ii) Hybrid masking: Feature-level masking is applied based on intervention probabilities and gradient-informed importance derived from GNN representations. iii) Graph structure modification: This involves altering edges to simulate alternative causal dependencies in the graph. The model maintains two separate GAT convolution pipelines to process contextual and object-related features, representing causal and non-causal entities respectively. This design enables the model to separate the causality aspect in the graph data.

\subsubsection{Mutual Information Estimation}

A dual-encoder mutual information estimator inspired by \cite{belghazi2018mutual} is employed to estimate and minimize the mutual information between the causal and non-causal components of node representations. These components are extracted from node embeddings obtained from separate GAT-based encoders. The estimator measures mutual information by contrasting positive pairs with negative samples using a similarity-based scoring function, thereby facilitating disentanglement by reducing informational overlap between the two views. 

For the causal encoder ($f_c$) and the spurious encoder ($f_o$), the objective function for mutual information minimization is defined in Eq.~\ref{eq-MI1}, where $T(.,.)$ denotes a similarity function and $Q$ represents a queue of negative samples. This is further refined through conditional MI estimation between $x_c$ and $x_o$, which reinforces the independence of the representations. To ensure a clear separation between the causal and non-causal subspaces, a cosine orthogonality constraint is imposed. Additionally, a learnable fusion gate dynamically combines the two components, with confidence-based supervision guiding the gate's weights to align with the predictive reliability of each pathway.

\begin{equation}\label{eq-MI1}
\mathcal{L}_{\text{MI}}(f_c, f_o) =
\mathbb{E}\left[ T(f_c, f_o) \right]
- \log \left( \mathbb{E}_{f_c' \in Q} \left[ e^{T(f_c', f_o)} \right] \right)
\end{equation}

\subsubsection{Interventional Training for Causal Learning}
The model undergoes interventional training through counterfactual feature interventions and multi-branch learning. Counterfactuals are generated by shuffling causal features and recombining them with original non-causal features through a learnable gate as defined in Eq.~\ref{eq-CF1}. Here, $\hat{y}_{\text{int}}$ denotes the predicted interventional output, $\alpha$ is a learnable gate coefficient, $\mathbf{x}_c^{\text{shuf}}$ represents the shuffled causal features, $\mathbf{x}_o$ represents the original non-causal features and $f_{\text{cls}}$ is the classifier function. Four prediction branches are used: causal, non-causal, fusion and intervention-based pathways. The model is trained using a combined loss that includes contrastive, center, and adaptive terms to better capture causal patterns and enhance robustness.

\begin{equation} \label{eq-CF1}
\hat{y}_{\text{int}} = f_{\text{cls}}\left( \alpha \cdot \mathbf{x}_c^{\text{shuf}} + (1 - \alpha) \cdot \mathbf{x}_o \right)
\end{equation}

\section{Experiments}
In this section, we present the experiment design, results and analysis used to evaluate the CCAGNN model.
\subsection{Experiment Design}
This work aims to explore the following research questions:
\begin{enumerate}
    \item \textbf{RQ1.} What impact do interventions on variables have in controlling confounding bias in causal classification tasks?
    \item \textbf{RQ2.} What effect does identifying key causal variables have on the accuracy and robustness of classification models in complex systems?
    \item \textbf{RQ3}. How do variations in graph connectivity influence the generalization performance of classification frameworks on new, similar graphs?
\end{enumerate}

\subsubsection*{Experimental Settings}
Experiments were carried out using Visual Studio Code 1.84 on Ubuntu 22.04, with Python 3.11 and PyTorch, using an NVIDIA GeForce RTX 3070 Ti GPU. The model was trained over 20 epochs with a learning rate of \textit{1e-3}, employing early stopping to prevent overfitting.

\subsection{Datasets}
We employ datasets from diverse domains including citation, co-purchase and social network graphs as discussed below.
\renewcommand{\arraystretch}{1.3}
 \begin{table}[h]

\caption{Summary of datasets used in the study.}

 \centering
{\fontsize{10}{10}\selectfont
  \begin{tabular}
  {|p{3cm}|p{1.3cm}|p{1.3cm}|p{1.5cm}|}
  \hline
      \textbf{Dataset}   & \textbf{ \# Nodes } &  \textbf{\# Edges } & \textbf{\# Classes} \\
   
\hline
Cora  &  2708  & 10556  & 7 \\
Citeseer    & 3327  &  9104 & 6 \\
PubMed  &  19717 & 88648  & 3 \\
Twitch    & 9498  &  315774 & 2 \\
Amazon-Computers  &  13752  &  491722 & 10 \\
Amazon-Photo    &  7650 & 238162  & 8 \\ 
\hline 
\end{tabular}
}
\label{tab:tabldatasets}
\end{table}
\textbf{Citation:} \textit{Cora}, \textit{Citeseer} and \textit{PubMed} are citation network datasets that categorize scientific publications into 7, 6 and 3 classes respectively. Nodes represent papers, and edges represent citation relationships, highlighting their significance in graph classification research across numerous studies \cite{velivckovic2017graph, kipf2016semi, wu2024graph}.  \\
\textbf{Amazon:} \textit{Amazon Computers} and \textit{Amazon Photo} are from the Amazon co-purchase graph, where nodes represent products and edges indicate that two products are often co-purchased. The labels represent product categories. These datasets have been used for graph classification tasks in previous studies including \cite{nasari2024enhancing, wang2024uncovering}. \\
\textbf{Twitch:} This is a social network dataset, where nodes represent Twitch streamers and edges represent social connections between them, and labels indicate streamer categories. The dataset contains various subgraphs and we use the \textit{DE} subgraph in our study. It has been used in related work such as \cite{wu2024graph}.

\renewcommand{\arraystretch}{1.3}
\begin{table*}[h]

\fontsize{9}{9}\selectfont
\caption{F-Scores (\%) for node classification tasks \label{tab:tableResults}}
  \centering
  {\fontsize{10}{10}\selectfont
   \begin{tabular}  {p{2.8cm} p{2.4cm} p{1cm} p{1.5cm} p{1.6cm} p{1cm} p{1cm} p{1cm}}
  \hline

   Category & \textbf{Model} &   Twitch  & Amazon-c & Amazon-p & Cora & Citeseer & PubMed  \\
    \hline
    
   Vanilla & GCN \cite{kipf2016semi}  &  96.39 &  97.78 &  98.85 & 72.61 & 44.88 & 40.00  \\

    methods & GAT \cite{velivckovic2017graph}  &  96.82 &  97.83 &  99.07 &  72.41 &  45.74 & 36.80  \\

    & GraphSAGE \cite{hamilton2017inductive}  &  74.36 &  91.24 &  96.81 &  70.81 & 42.07 & 39.01 \\ 
      \noalign{\hrule height 0.1pt}
    
      & GCN-CAL \cite{sui2022causal} &  97.14 &  98.32 &  99.46 &  68.80 &  39.67 & 76.84  \\
     
     Causal & GAT-CAL \cite{sui2022causal}  & 98.10 &  98.22 &  99.34 &   66.27 &  39.06 & 85.42\\   

    methods & HebCGNN \cite{job2025hebcgnn} &  98.30 &  98.21 &  99.34 & 75.63  &  58.18 & 64.30 \\
    
    & GCN-ICL \cite{zhao2024twist} &  95.98 &  98.36 &  99.41 &   73.94 &  42.71 & 77.91  \\

    & GAT-ICL \cite{zhao2024twist} &  96.71 &  98.31 &  98.50 &   74.47 &  51.76 & 84.15  \\
    \noalign{\hrule height 0.1pt}

      & ACE-GCN \cite{chen2025revolutionizing} & 64.50 & 27.65 &  26.00 &  61.90 &  35.93 & 73.03 \\ 
         Intervention-based & ACE-GAT \cite{chen2025revolutionizing} &  68.12 &  27.06 &  26.95 &  72.37 &  58.88  & 78.67\\  

         causal methods   & CaNet \cite{wu2024graph} &  66.17 &  09.31 & 51.30 &  84.32 &   73.70 &  87.19\\ 
     & \textbf{CCAGNN} (ours)  &  \textbf{98.93} & \textbf{98.76} &  \textbf{99.83} &   \textbf{91.14} &  \textbf{82.84} & \textbf{89.97}  \\ 
    \hline

    \hline
    
\end{tabular}
}
\end{table*}

\subsection{Baseline Models}

Vanilla Methods: 
\begin{itemize}
  \item GCN \cite{kipf2016semi}: Graph Convolutional Networks (GCN) is the standard GNN architecture used to assess performance improvements in enhanced variants. GCN utilizes a convolutional mechanism to propagate information across a graph through feature aggregation. 
    
    \item GAT \cite{velivckovic2017graph}: GAT (Graph Attention Network), a variant of GNNs, uses an attention mechanism to assign weights to neighbouring node information. This enables the model to capture complex dependencies, thus making it effective for graph classification.

    \item GraphSAGE \cite{hamilton2017inductive}: GraphSAGE (Graph Sample and Aggregation) learns node representations by sampling and aggregation of neighbourhood features. The node embeddings are iteratively updated using this aggregated information.
\end{itemize}   
Causal Methods: 
\begin{itemize}
    \item GCN-CAL, GAT-CAL \cite{sui2022causal}: The CAL framework explored causality for GNN classification. The model primarily used a GNN-based encoder for obtaining node representations, followed by two linear layers for edge and node-level attention. The framework employed GCN (GCN-CAL) and GAT (GAT-CAL) architectures for causal classification. 
    \item HebCGNN  \cite{job2025hebcgnn}: HebCGNN is a Hebbian-enabled causal GNN that uses dynamic impact valuing to highlight causal relationships in complex graph features, improving classification over traditional methods.
    \item GCN-ICL, GAT ICL \cite{zhao2024twist}: The Information-based Causal Learning (ICL) framework integrates information theory and causality to enhance GNNs by prioritizing causal features over correlational ones. 
\end{itemize}
Intervention-based Causal Methods: 

\begin{itemize}
\item CaNet \cite{wu2024graph}: Causal Intervention for
 Network Data (CaNet) is a causal framework that uses an environment estimator and mixture-of-expert GNN to counteract latent confounding bias, improving node-level OOD generalization without needing environment labels.
 \item ACE \cite{chen2025revolutionizing}: The Adaptive Causal Module and Causality-Enhanced (ACE) framework extracts causal features and mitigates spurious correlations using an adaptive causal module, controlled backdoor adjustment and a causality enhancement module, enabling improved generalization without additional training.
\end{itemize}

\subsection{Experiment Results}

In this section we present and analyse the experimental results. The classification F-scores are provided in Tab.~\ref{tab:tableResults}. \textit{CCAGNN} and all baseline models were trained for 20 epochs, consistent with the work of \cite{job2025hebcgnn}. An exception is the ACE framework, which showed F-scores below 20 for all datasets except \textit{Twitch} and \textit{PubMed}; therefore we trained it for 80 epochs to allow for a reasonable comparison. In addition to the F-scores, the Mutual Information (MI) loss over the training epochs are plotted in Fig.~\ref{fig:causalMIloss} to monitor the disentanglement  progress.

\subsubsection*{Performance Metrics}
The model is assessed using 5-fold cross validation, with the F-score as the evaluation metric. The F-score is well-suited for node classification tasks because it balances precision and recall, which is especially important when handling imbalanced class distributions, which is a common issue in graph-based data. 

\subsection{Analysis of Results}
In this section, we analyse the results presented in Tab.~\ref{tab:tableResults}. We analyse the baselines under three categories (i) vanilla methods, including the basic GCN, GAT and GraphSAGE architectures, where no causal modeling is integrated; (ii) causal methods such as the CAL framework \cite{sui2022causal}, HebCGNN \cite{job2025hebcgnn} and the ICL framework \cite{zhao2024twist} and (iii) intervention-based causal methods, including the ACE framework \cite{chen2025revolutionizing} and CaNet \cite{wu2024graph}. 

Vanilla methods show strong performance on the \textit{Amazon} and \textit{Twitch} datasets, with GCN and GAT achieving F-scores above 96\%. However, GraphSAGE under-performs slightly on \textit{Amazon-Computers} (91.24\%) and much lower on \textit{Twitch} (74.36\%). All three vanilla methods exhibit a significant drop in performance on the citation datasets, particularly \textit{Citeseer} and \textit{PubMed}, likely due to their limited ability to model complex dependencies and underlying confounding in node interactions.

The CAL variants (causal methods) present slight improvements over  vanilla models on the \textit{Amazon} and \textit{Twitch} datasets. Their performance on citation datasets is mixed; with GAT-CAL performing well on \textit{PubMed} (85.42\%), and both CAL variants performing poorly on \textit{Citeseer} with F-scores around 39\%. HebCGNN shows more consistent results across datasets, though the performance on \textit{PubMed} is lower at 64.30\%. On the other hand, ICL-based models outperform CAL in most cases and demonstrate better generalization across datasets. GAT-ICL, in particular, yields good F-scores of 74.47\% (\textit{Cora}), 51.76\% (\textit{Citeseer}) and 84.15\% (PubMed) for the citation datasets. Compared to vanilla models, causal methods, specifically HebCGNN and ICL demonstrate more stable performance across the datasets, emphasizing the benefits of explicitly addressing confounding and spurious correlations.

Intervention-based causal models, ACE and CaNet aim to isolate causal effects through explicit interventions. Both ACE variants perform poorly on the \textit{Amazon} datasets (around 26\%), and moderately on \textit{Twitch} datasets (64.50\% and 68.12\%), suggesting that their intervention strategies may be less effective or potentially over corrected in large-scale graphs. Notably, ACE-GAT performs reasonably well on the citation datasets compared to vanilla models. CaNet performs better than ACE variants on the citation datasets, with notably high scores on all three datasets: Cora (84.32\%), Citeseer (73.70\%) and PubMed (87.19\%). Despite their ability to handle confounding, the overall inconsistency of intervention-based models highlights the difficulty of designing effective interventions in complex, real-world graph settings.

Our proposed model, \textit{CCAGNN} achieves the highest performance across all six datasets, with substantial margins observed on the citation datasets: \textit{Cora}: 91.14\%, \textit{Citeseer}: 82.84\% and \textit{PubMed}: 89.97\%. These results indicate that \textit{CCAGNN} effectively integrates causal reasoning into graph learning and is robust to confounding, selection bias and distribution shifts, which are factors that commonly affect performance in standard GNNs.

To summarise, vanilla models perform well on datasets like \textit{Twitch} and \textit{Amazon}, where node features align well with labels. Causal methods improve performance on more challenging datasets but vary in their effectiveness. Intervention-based approaches show potential but lack consistency. \textit{CCAGNN} consistently outperforms all baselines and state-of-the-art causal models, demonstrating the advantage of combining causal insights with graph neural architectures.

\begin{figure}%
    \centering
    \subfloat[\centering \textit{Cora} dataset]{{\includegraphics[width=7cm]{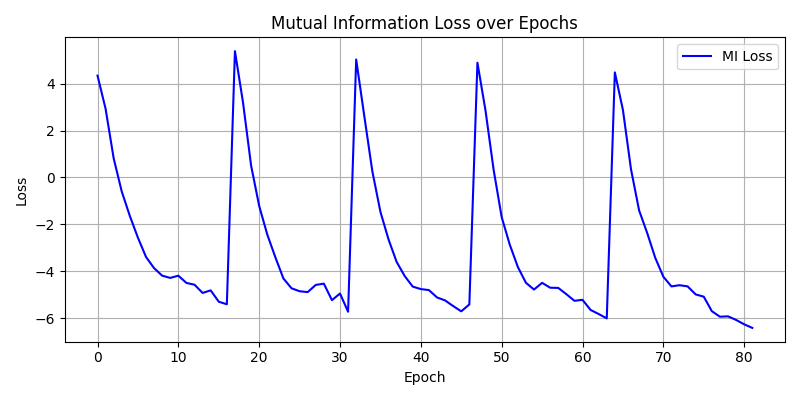} }}%
    \qquad
    \subfloat[\centering \textit{Amazon-Computers} dataset]{{\includegraphics[width=7cm]{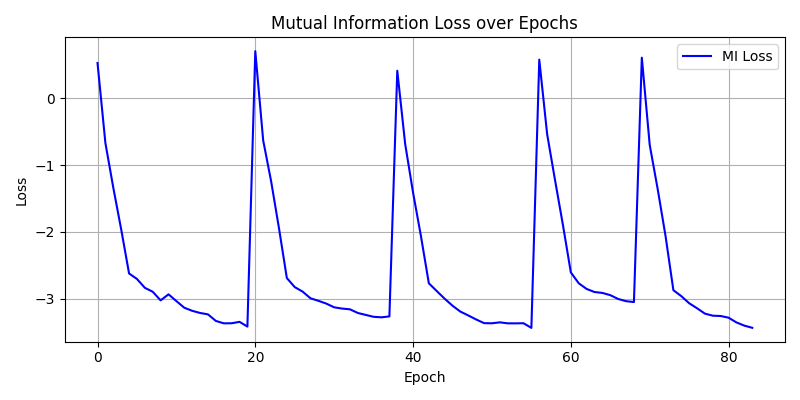} }}%
   \caption{Mutual Information Loss Plots}
    \label{fig:causalMIloss}
\end{figure}

In addition to the F-score results, we monitor the disentanglement  progress to evaluate the model's ability to reduce mutual information between the causal and non-causal pathways, thereby mitigating the impact of confounding variables. The MI loss for two datasets, one from the citation category and the other from a general category) is shown in Fig.~\ref{fig:causalMIloss}. Each plot represents training across five folds, where the beginning of each fold is indicated by a peak in the curves. For the \textit{Cora} dataset, the MI loss exhibits a sharp drop early in training, indicating that the model quickly begins to disentangle causal and non-causal factors by reducing their dependency. This indicates effective learning of disentangled representations. In later epochs, the model enters a fine-tuning phase where it maintains performance while enforcing disentanglement. Towards the end of training, the MI loss further reduces, which indicates a strong separation, that is minimal mutual information between the causal and non-causal pathways. A similar trend is observed in the \textit{Amazon-Computers} dataset, although the MI loss starts at a lower value than in \textit{Cora}. In both datasets, the model successfully reduces MI  over time, with most of the learning occurring in the first half of training, with the latter half contributing to stabilization.

\subsubsection*{Ablation Studies}

\begin{figure}[h] 
\centering

\includegraphics[width=0.75\textwidth]{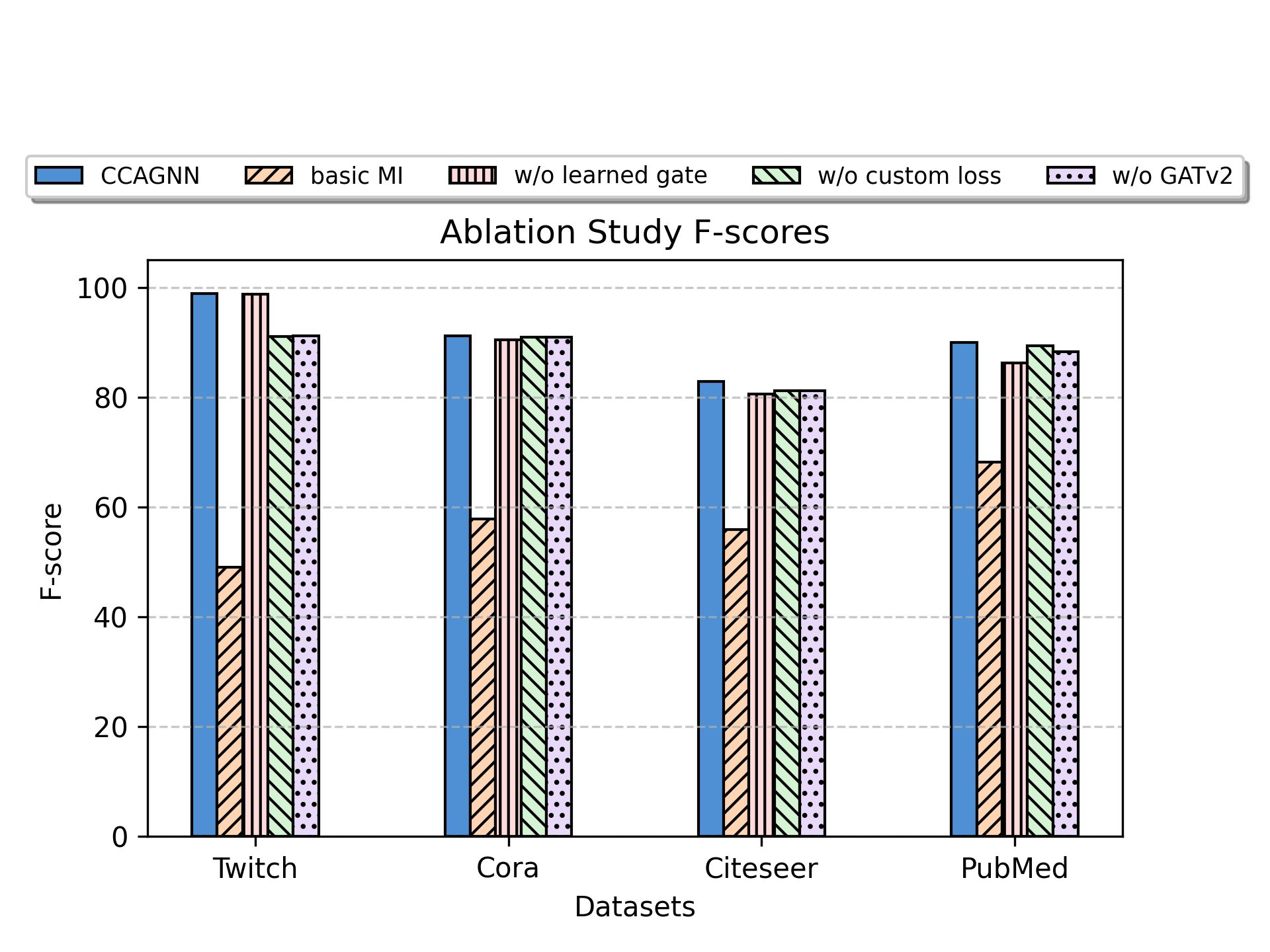}
\caption{Ablation study results showing F-scores for different model configurations across datasets.}
\label{fig:ablation}
\end{figure}

We conducted an ablation study to evaluate the contribution of each component of the CCAGNN model to its overall performance. We studied four configurations and the F-score results are shown in Fig.~\ref{fig:ablation}. First, we used a basic form of mutual information model with a single encoder, referred to as \textit{basic MI} in the figure. Second, we evaluated the model without the learned gate mechanism (\textit{w/o learned gate}). Third, we removed the custom losses described in the methodology section (\textit{w/o custom loss}) and lastly we replaced the GATv2 layers with basic GAT layers to assess their impact. As shown in the figure, CCAGNN achieves the highest F-scores across all four datasets, highlighting the effectiveness of the overall framework. The most substantial performance drops are observed with the \textit{basic MI} variant across all datasets, and with \textit{w/o custom loss} and \textit{w/o GATv2} on the \textit{Twitch} dataset. There is a considerable drop in F-scores across all four ablated variants for the \textit{Citeseer} dataset, as well in the \textit{w/o learned gate} and \textit{w/o GATv2} settings on the \textit{PubMed} dataset. These results highlight the importance of all four components in the CCAGNN framework, with mutual information enhancement being most significant.

\subsubsection*{Summary of Results}

Based on these results, we address the research questions as follows:

\begin{enumerate}
    \item \textbf{RQ1.} As an intervention-based causal model, \textit{CCAGNN} demonstrates that targeted interventions can significantly reduce the influence of confounding bias in node classification tasks. By applying controlled interventions on variables during the learning process, \textit{CCAGNN} isolates causal effects from spurious correlations. This results in significant improvements in performance, particularly on structurally complex datasets like Citeseer (82.84\%) and PubMed (89.97\%), where traditional as well as other causal models perform inconsistently. The results confirm that well-designed, model-integrated interventions can effectively enhance learning stability in causal graph settings. Additionally, the MI loss plot demonstrates that interventions promote disentanglement, thereby helping to mitigate confounding effects.
    
    \item \textbf{RQ2.} \textit{CCAGNN} uses interventions to identify and incorporate key causal variables into the learning process, separating them from irrelevant or confounding features. This enables the model to focus on causally meaningful information, rather than spurious patterns that may vary across domains or graph structures. The model’s consistent performance across diverse datasets highlights the importance of causal variable identification in building robust and accurate classifiers. The evidence suggests that interventions aid not only in debiasing but also in enhancing model generalizability by promoting causal consistency.
    
    \item \textbf{RQ3}. \textit{CCAGNN} shows strong generalization performance across datasets with varying graph connectivity, suggesting that its intervention-based causal reasoning reduces reliance on node attribute similarity and structural uniformity. Though traditional GNNs under-perform significantly on graphs with diverse labels and sparse connections, \textit{CCAGNN} maintains high accuracy. This indicates that its causal intervention mechanism enables it to learn structure-independent, stable representations, making it less affected by connectivity variations and more reliable in real-world settings where graph structure is not consistent.
    
\end{enumerate}

\section{Conclusion}
In this paper, we present \textit{CCAGNN}, a confounder-aware causal graph neural network designed for graph classification tasks, which integrates causality and counterfactual analysis to enable accurate, causally grounded predictions that generalize well to real-world scenarios. The model captures causal dependencies in graph data by applying targeted interventions and counterfactual reasoning to simulate feature changes. Causality is integrated into the attention-based GNN through interventional message passing and contrastive training with mutual information regularization. Experimental results on six public datasets show that \textit{CCAGNN} outperforms existing state-of-the-art approaches. Future work will explore dynamic causal modeling and further enhance the interpretability of causal graph neural networks models.

\begin{acks}
This work is partially supported by grants from Australian Research Council (No. DP220101360) and the SAGE Athena Swan Scholarship, UniSQ.
\end{acks}

\bibliographystyle{ACM-Reference-Format}
\bibliography{sj-ccagnn}


\end{document}